\documentclass[10pt,twocolumn,letterpaper]{article}

\usepackage{cvpr}              

\usepackage{graphicx}
\usepackage{amsmath}
\usepackage{amssymb}
\usepackage{booktabs}
\usepackage{multirow}

\usepackage[pagebackref,breaklinks,colorlinks]{hyperref}

\usepackage[capitalize]{cleveref}
\crefname{section}{Sec.}{Secs.}
\Crefname{section}{Section}{Sections}
\Crefname{table}{Table}{Tables}

\def\cvprPaperID{*****} 
\def\confName{CVPR}
\def\confYear{2022}

\begin{document}

\title{Advancements in Repetitive Action Counting: Joint-Based PoseRAC Model With Improved Performance}

\author{Haodong Chen, Niloofar Zendehdel, Ming C. Leu\\
Department of Mechanical and Aerospace Engineering\\
Missouri University of Science and Technology\\
{\tt\small h.chen@mst.edu}\\
\and
Md Moniruzzaman, Zhaozheng Yin\\
Department of Biomedical Informatics and Department of Computer Science\\
Stony Brook University\\
\and
Solmaz Hajmohammadi\\
Peloton Interactive\\
}
\maketitle
\begin{abstract}
Repetitive counting (RepCount) is critical in various applications, such as fitness tracking and rehabilitation. Previous methods have relied on the estimation of red-green-and-blue (RGB) frames and body pose landmarks to identify the number of action repetitions, but these methods suffer from a number of issues, including the inability to stably handle changes in camera viewpoints, over-counting, under-counting, difficulty in distinguishing between sub-actions, inaccuracy in recognizing salient poses, etc. In this paper, based on the work done by \cite{yao2023poserac}, we integrate joint angles with body pose landmarks to address these challenges and achieve better results than the state-of-the-art RepCount methods, with a Mean Absolute Error (MAE) of 0.211 and an Off-By-One (OBO) counting accuracy of 0.599 on the \textit{RepCount} data set \cite{hu2022transrac}. Comprehensive experimental results demonstrate the effectiveness and robustness of our method. 

\textbf{keywords}:Repetitive action counting, \textit{RepCount}, Video Transformer, Skeleton, Pose landmarks

\end{abstract}

\section{Introduction}
\label{sec:intro}
Repetitive counting (RepCount) is essential for monitoring physical activity, assessing exercise performance, and guiding the rehabilitation process. Accurate and reliable RepCount algorithms are critical for providing accurate feedback to users, enabling objective assessments, and facilitating personalized fitness programs. Similarly, in scientific experimentation, the precision and repeatability of actions are paramount, as they directly influence the validity and reliability of experimental results. The ability to analyze and quantify repetitive actions becomes a crucial tool for ensuring correctness, efficiency, and quality across diverse contexts and applications~\cite{albu2008generic, chen2023real, gonzalez2021durable, briassouli2007extraction, hao2020filtration, chen2020design, brickwood2019consumer, foran2001high, hao2021factors, shen2017milift, chen2022real}.
Existing RepCount methods, such as the method in \cite{hu2022transrac}, rely on the red-green-and-blue (RGB) frame inputs, which provide a solution for RepCount but cannot independently isolate and recognize periodic movement. Some other studies, such as those by \cite{dwibedi2020counting} et al. and \cite{zhang2020context} et al., have achieved better performance by using contextual information in repetitive actions. In addition, \cite{yao2023poserac} proposed a pose saliency transformer for RepCount, which achieved state-of-the-art performance that has been maintained until our improvement. In this paper, we integrate joint angles with the pose landmark model proposed by \cite{yao2023poserac} to address the RepCount problem, and our method addresses several existing issues in RepCount, including the inability to stably deal with changes in camera viewpoints, over-counting, under-counting, difficulty in distinguishing between sub-actions, and  inaccuracy in recognizing salient poses. By addressing these issues, our model is able to perform a more accurate and robust counting of repetitive actions on the \textit{RepCount} data set proposed by \cite{hu2022transrac}.

\subsection{Related Works}
\begin{figure}[h]
    \centering
    \includegraphics[width=\linewidth]{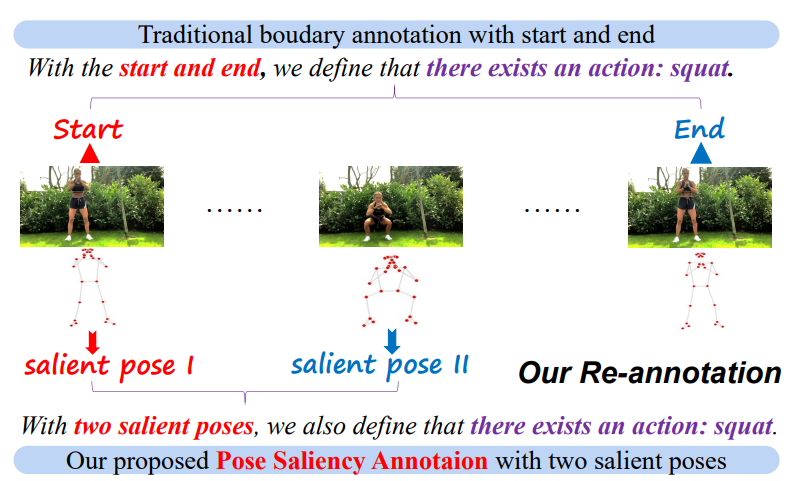}
    \caption{Pose saliency annotation. Instead of annotating the
start and end, the two most salient poses were annotated in \cite{yao2023poserac}}
    \label{fig:sal_anno}
\end{figure}

In our work, we adopt and improve the RepCount method proposed in \cite{yao2023poserac}, which introduces a new mechanism called Pose Saliency Representation (PSR). This mechanism uses the two most salient poses to represent the action, providing a more efficient alternative to the common RGB frame-based representation. Traditional RGB frame-based methods require complex computations to obtain high-level semantic information from intra-frame spatial and inter-frame temporal information. In contrast, as shown in \cref{fig:sal_anno}, the PSR utilizes only two salient poses to capture the features of each action, thus establishing a unique mapping between salient poses and RepCount.

Based on the PSR, a Pose Saliency Transformer for Repetitive Action Counting (PoseRAC) is proposed in \cite{yao2023poserac}. Their PoseRAC consists of three key components: (1) pose extraction, which uses state-of-the-art algorithms, i.e., BlazePose \cite{bazarevsky2020blazepose}, to extract poses from all frames; (2) a core model, which maps each extracted pose to an action category; and (3) a lightweight action trigger model, which performs video-level RepCount using the actions in each frame.

In work done by \cite{yao2023poserac}, RepCount relies heavily on human pose estimation, which uses landmark detection to identify the joint positions of human skeletons. However, these methods often suffer from several drawbacks, such as the inability to stably deal with changes in camera viewpoints, over-counting, under-counting, difficulty in distinguishing between sub-actions, inaccuracy in recognizing salient poses, etc.

To overcome these limitations, we propose a novel approach that integrates 5 joint angles with body poses for the RepCount analysis, which improves performance and addresses the aforementioned challenges. 

\subsection{Contribution}
The contribution of this paper is as follows:
\begin{itemize}
    \item We analyze different combinations of joint angles and body pose landmarks in solving the RepCount problem.
    \item We improve the RepCount performance in dealing with changes in camera viewpoints, over-counting, under-counting, difficulty in distinguishing between sub-actions, inaccuracy in recognizing salient poses, etc.
    \item Our experimental results obtain better performance than the state-of-the-art results on the public data set (\textit{RepCount}). \cite{hu2022transrac}.
\end{itemize}
The structure of the rest content is as follows: we first introduce our approach in Section \cref{sec:approach}. Next, we provide experimental results in Section \cref{sec:results}. Finally, we give conclusions in Section \cref{sec:conclusion}.

\section{Pose and Joint Angle Annotation}
\label{sec:approach}
\begin{figure}[h]
    \centering
    \includegraphics[width=0.8\linewidth]{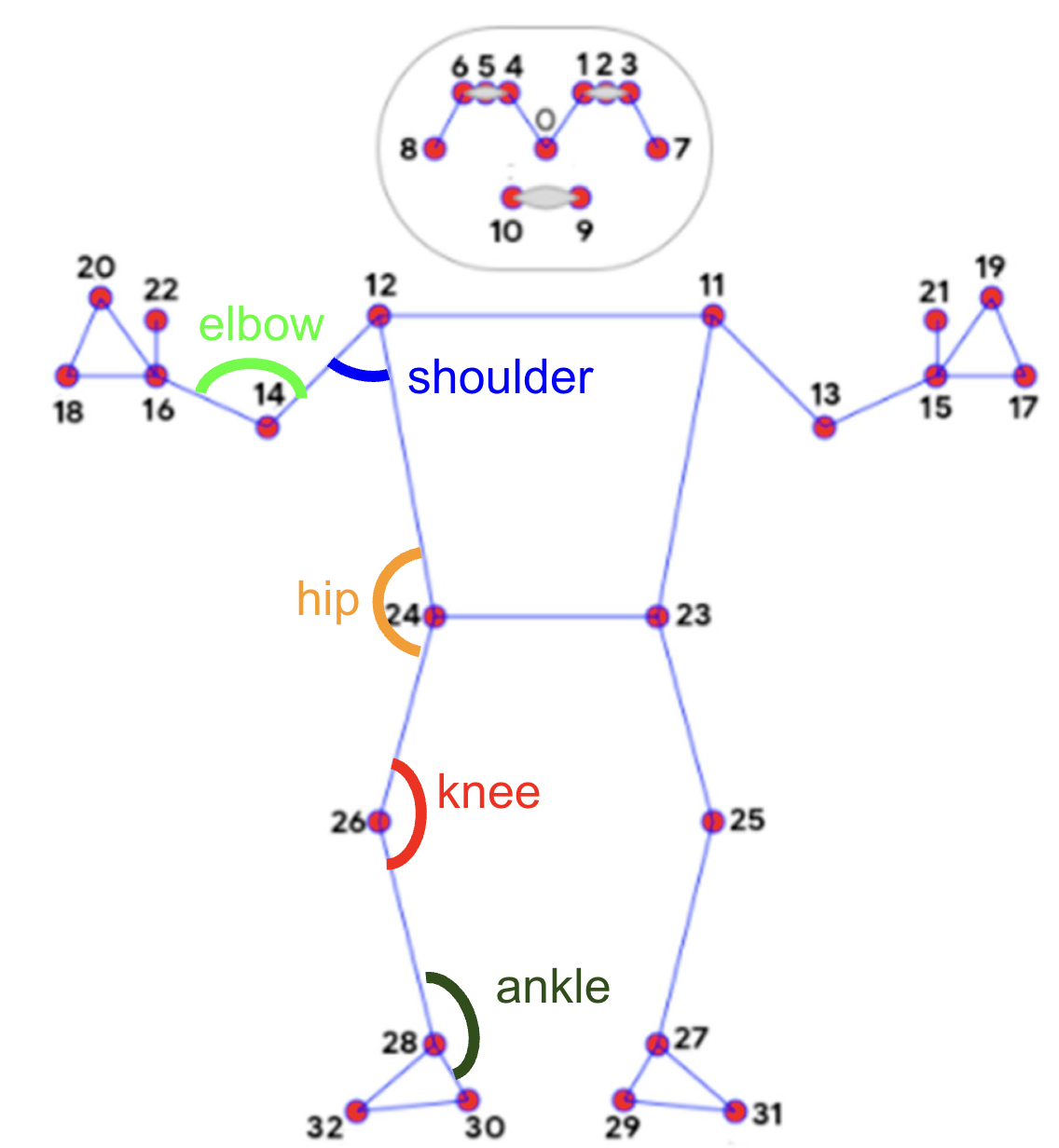}
    \caption{BlazePose landmarks and five joint angles}
    \label{fig:landmarks}
\end{figure}

As shown in \cref{fig:landmarks}, the 33 pose landmarks are extracted by the Google Mediapipe BlazePose model \cite{bazarevsky2020blazepose, chen2023fine}. Five joint angles are extracted based on the landmarks shown in \cref{fig:landmarks}, which are elbow, shoulder, hip, knee, and ankle angles. 

\subsection{Data Set Annotation Correction}

After analyzing the original data set \textit{RepCount} \cite{hu2022transrac}, we noticed that the ground truth of the data stu4\_5.mp4 in the test data set was incorrectly labeled as 51, whereas the correct ground truth label should be 5, and we corrected the label annotation.

\subsection{RepCount Visualization - Density Map}
\begin{figure}[h]
    \centering
    \includegraphics[width=\linewidth]{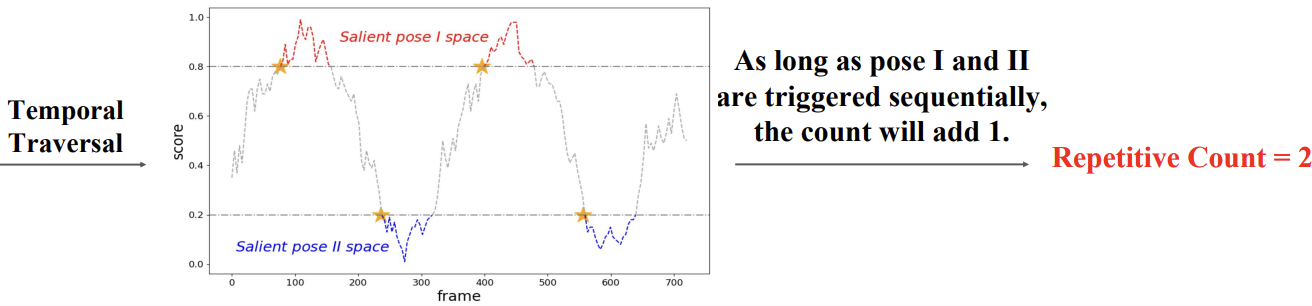}
    \caption{Action-trigger mechanism using density map \cite{yao2023poserac}. }
    \label{fig:action-trigger}
\end{figure}

After pose mapping using the Swin Transformer in \cite{liu2022video}, we can obtain the score for each frame and generate a density map as shown in \cref{fig:action-trigger}. Higher values indicate a higher similarity to the salient pose I, while lower values indicate a higher match to the salient pose II. The action-trigger mechanism is used to compute the time at which two salient poses appear in sequence in an action category, where a specific upper and lower limit is set to differentiate the scores of the two salient poses, thus clustering the non-salient poses in the middle and easily categorizing the salient poses at both ends \cite{yao2023poserac, onoro2016towards, sreenu2019intelligent}.

\section{Experiments and Results}
\label{sec:results}
\subsection{Experiment Setup}
The experimental platform is a workstation with an Ubuntu 16.04 system equipped with an Intel Xeon Gold 6226R CPU, an NVIDIA GeForce RTX 3090 graphics card, and 64343M RAM. The model was trained on one GPU in 20 minutes for  20 epochs.

\subsection{Evaluation of Different Scenarios Using Joint Angles}
We evaluate different scenarios using joint angles in solving the RepCount problem, including:
\begin{itemize}
    \item using the five left joint angles, i.e., [left $A_e$, left $A_s$, left $A_h$, left $A_k$, left $A_a$ ]
    \item using five left and right joint angles, i.e.,  [left $A_e$, right $A_e$, left $A_s$, right $A_s$, left $A_h$, right $A_h$, left $A_k$, right $A_k$, left $A_a$, right $A_a$]
    \item using average values of the five left and right joint angles, i.e., [average $A_e$, average $A_s$, average $A_h$, average $A_k$, average $A_a$]
\end{itemize} 

The $A_e$, $A_s$, $A_h$, $A_k$, and $A_a$ represent the elbow, shoulder, hip, knee, and ankle angles, respectively. We consider the single-side joint angles because all actions in the \textit{RepCount} data set are symmetric. 

In this paper, we adopt the main evaluation metrics used in previous work \cite{dwibedi2020counting, zhang2020context, hu2022transrac, yao2023poserac}, i.e., mean absolute error (MAE) and off-by-one (OBO) counting accuracy. MAE represents the normalized absolute error between ground truth and prediction, while OBO measures the repetitive count rate of the entire data set. They can be defined as:

\begin{equation} 
\label{eu_OBO}
MAE = \frac{1}{N}\sum_{i=1}^{N} \frac{\left | \widetilde{c}_{i} - c_{i} \right |}{\widetilde{c}_{i}}
\end{equation}
\begin{equation} 
\label{eu_OBO}
OBO = \frac{1}{N}\sum_{i=1}^{N}\left [ \left | \widetilde{c}_{i} - c_{i} \right | \leq 1 \right ]
\end{equation}
where $\widetilde{c}_{i}$ is the ground truth count, $c_{i}$ is the prediction count, and $N$ is the number of videos.

The comparison results for the different scenarios are shown in \cref{tab:comparison:landmarks_angles}. We find that the case using landmarks and joint angles performed better than the case using only landmarks. In the cases using landmarks and joint angles, the best performance is obtained by using landmarks and the average values of the five left and right joint angles, in which the MAE $\approx$ 0.211 and OBO $\approx$ 0.599. \cref{tab:comparison:other_people} shows that on the \textit{RepCount} data set, our method consistently outperforms previous methods for both evaluation metrics, with an MAE metric of 0.211 compared to the 0.236 of the PoseRAC and an OBO metric of 0.599 compared to the 0.559 of the PoseRAC.
 

\begin{table*}[h]
    \centering
    \begin{tabular}{lll}
        \hline
        Different Cases                                                          & MAE   & OBO  \\ \hline
        Only landmarks     \cite{yao2023poserac}                                                      & 0.236 & 0.559 \\
        Landmarks + five left joint angles                                       & 0.227 & 0.571 \\
        Landmarks + five left and right joint angles                       & 0.213 & 0.587 \\
        \textbf{Landmarks + average values of the five left and right joint angles} & 0.211 & 0.599 \\ \hline
    \end{tabular}
    \caption{Comparison of different cases using landmarks and joint angles.}
    \label{tab:comparison:landmarks_angles}
\end{table*}

\begin{table*}[h]
\centering
    \begin{tabular}{llll}
    \hline
    \multicolumn{2}{l}{\multirow{2}{*}{Methods}}                          & \multicolumn{2}{l}{RepCount} \\ \cline{3-4} 
    \multicolumn{2}{l}{}                                                  & MAE           & OBO          \\ \hline
    \multirow{3}{*}{Video-level} & Zhang et al. \cite{zhang2020context}                        & 0.879         & 0.155        \\
                                 & Huang et al. \cite{huang2020improving}                     & 0.526         & 0.160        \\
                                 & TransRAC \cite{hu2022transrac}                            & 0.443         & 0.291        \\ \hline
    \multirow{2}{*}{Pose-level}  & PoseRAC \cite{yao2023poserac}                               & 0.236         & 0.559        \\ \cline{2-4} 
                                 & \textbf{PoseRAC + joint angles (Ours)} & 0.211         & 0.599        \\ \hline
    \end{tabular}
    \caption{Performance comparison on RepCount(-pose).}
    \label{tab:comparison:other_people}
\end{table*}

\subsection{Visualization Comparison of Models: Landmark-Only vs. Landmarks + Joint Angles}

To validate the effectiveness of our proposed method, we visually analyze the output density maps obtained using two models: the model using only the landmarks and the model integrating the 5 average joint angles with the landmarks. The density map represents the density of a particular human pose in an input video sample. Higher values (close to 1) indicate a higher similarity to salient pose I, while lower values (close to 0) indicate a higher match to salient pose II. The density map provides insight into the distribution of the two salient poses throughout the video. Our experiments focus on visualizing and comparing the following issues in RepCount: inability to stably deal with changes in camera viewpoints, over-counting, under-counting, difficulty in distinguishing sub-actions, inaccuracy in recognizing salient poses, etc.

\begin{figure}[h]
    \centering
    \includegraphics[width=\linewidth]{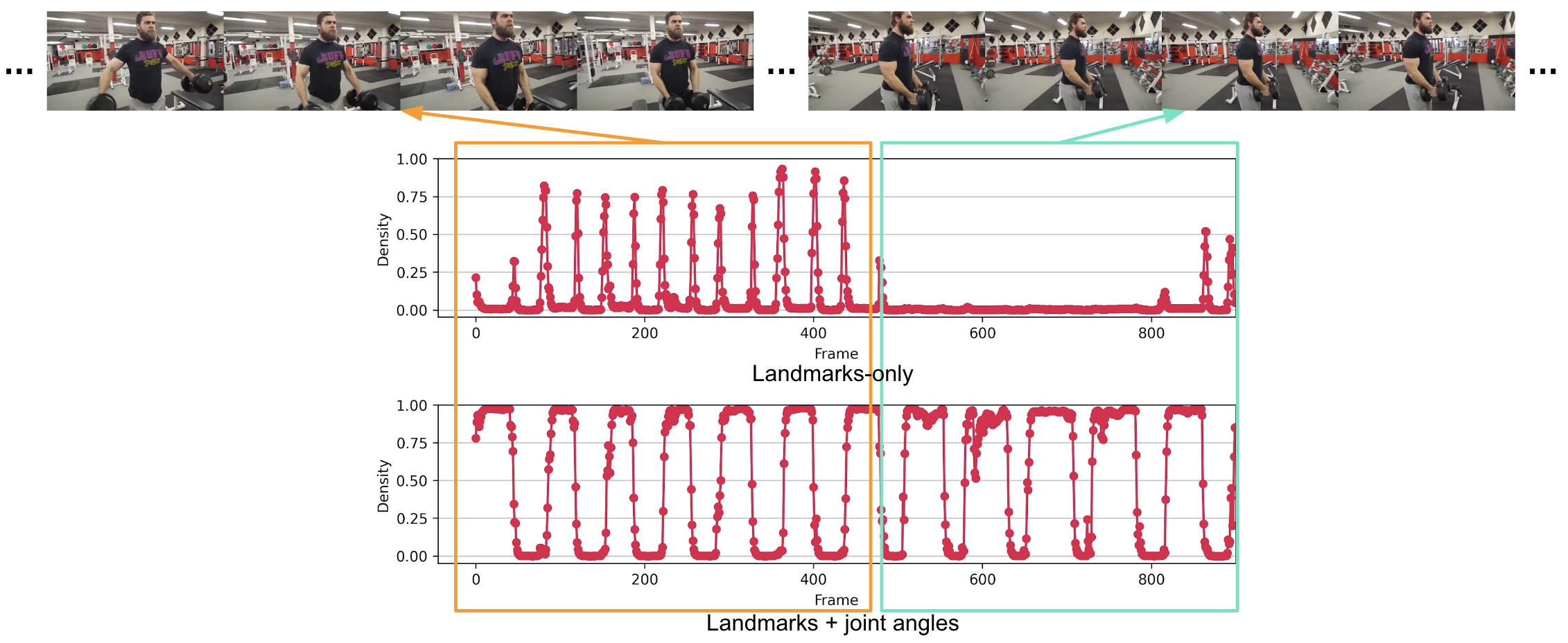}
    \caption{Density maps: landmarks-only vs. landmarks + joint angles - addressing the inability issue to stably deal with changes in camera viewpoints}
    \label{fig:density_camv}
\end{figure}

\textbf{Inability to stably deal with changes in camera viewpoints}: As shown in \cref{fig:density_camv}, we observe that the density map integrating the 5 average joint angles with the landmarks exhibits a more accurate capture of salient poses when the camera viewpoint changes. Specifically, the density values for the salient pose I remain consistently high (close to 1) after changing the camera viewpoint. In contrast, the density map using only the landmarks shows a significant drop in density values after the camera viewpoint changes from the front view to the side view, with density values $\approx$ 0 in the 440-800 frame range. This difference suggests that compared to using only the landmarks, using both the landmarks and 5 average joint angles enables the model not to miss certain salient poses during camera viewpoint changes, thereby maintaining the accuracy of RepCount.
  
\begin{figure}[h]
    \centering
    \includegraphics[width=\linewidth]{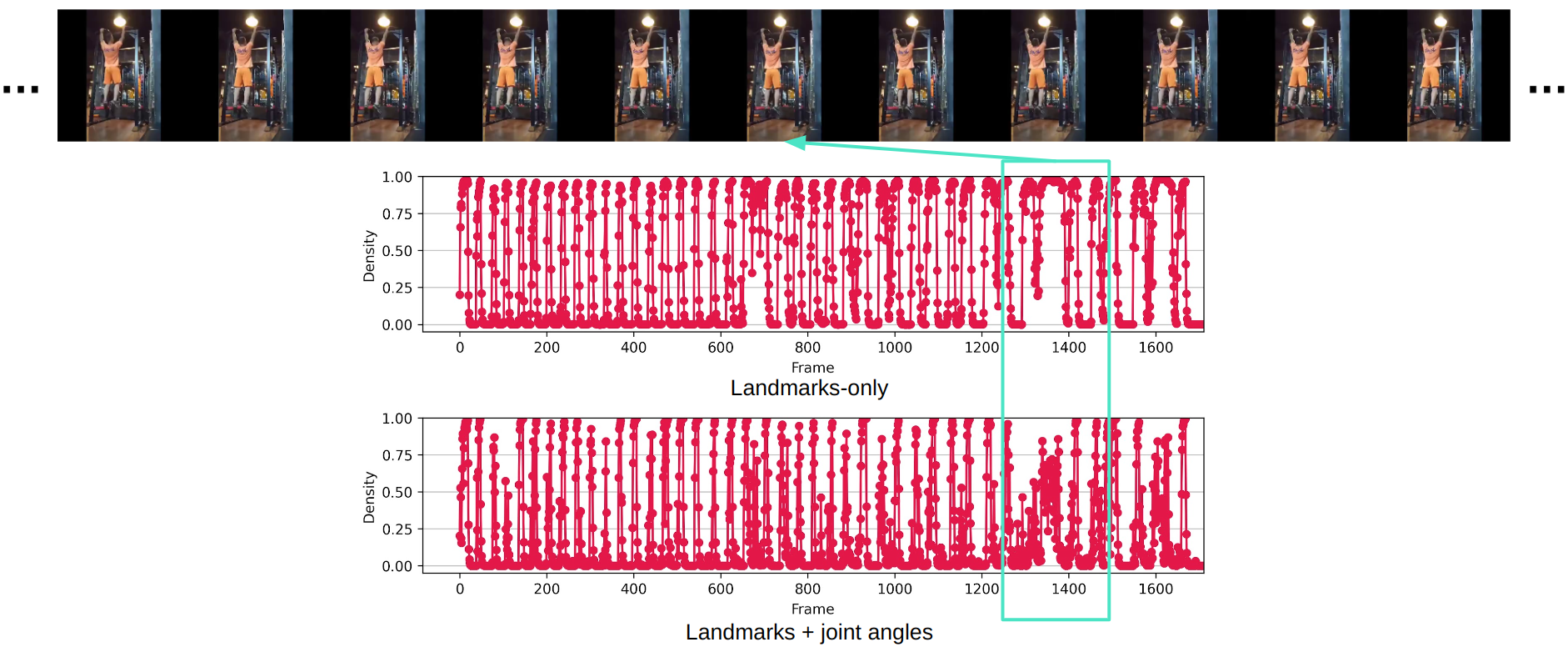}
    \caption{Density maps: landmarks-only vs. landmarks + joint angles - addressing the over-counting issue}
    \label{fig:density_oc}
\end{figure}

\textbf{Over-counting}: As shown in \cref{fig:density_oc}, the density map generated from the landmarks alone tends to produce over-counts when a subject attempts to perform the \textit{Pull Up} action but fails to complete it due to fatigue. The over counts are generated due to the subject's slight movement around the salient pose where the subject's arms are extended. In such cases, a landmark-only model may misinterpret these attempts as valid transitions between two salient poses, resulting in overcounts in RepCount. However, incorporating the 5 average joint angles allows the model to recognize salient poses more accurately. As shown in \cref{fig:density_oc}, when the subject attempts to perform a \textit{Pull Up} action but fails to complete it, this phenomenon is reflected in the density map obtained by integrating the 5 average joint angles with the landmarks, where the density value is lower than that for the salient pose I, i.e., the subject fully completes the \textit{Pull Up} action with the arms bent. Integrating the 5 average joint angles helps the model recognize when a subject's attempts are unsuccessful, thus avoiding over-counting these partial or incomplete attempts.

\begin{figure}[t!]
    \centering
    \includegraphics[width=\linewidth]{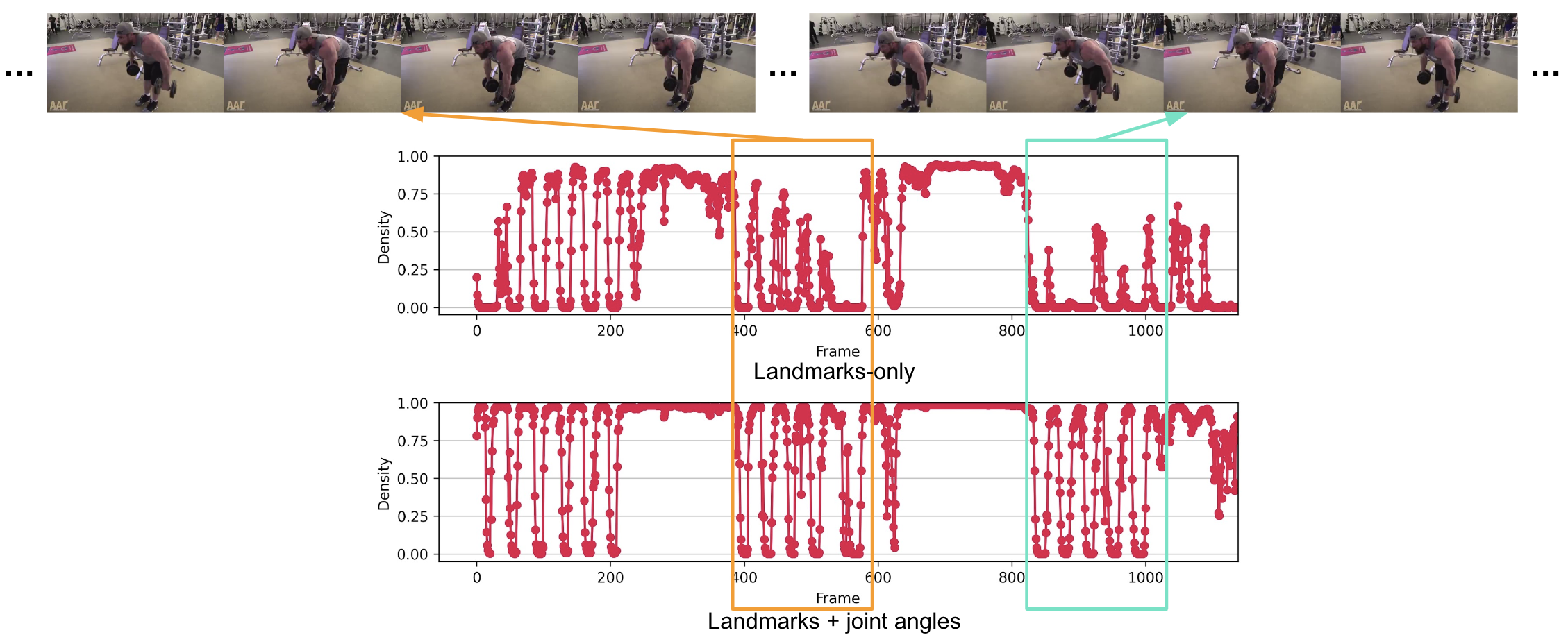}
    \caption{Density maps: landmarks-only vs. landmarks + joint angles - addressing the under-counting issue}
    \label{fig:density_uc}
\end{figure}

\textbf{Under-counting}: As shown in \cref{fig:density_uc}, when a subject endeavors to perform a \textit{Side Raise} action continuously, the density map derived exclusively from the landmarks fluctuates irregularly between 0 and 1 and shows multiple peaks between 0.5 and 0.75, rather than regularly fluctuating between 0 (salient posture II) and 1 (salient posture I). However, the density map integrating the 5 average joint angles with the landmarks shows density values regularly fluctuating between 0 and 1, representing the subject's movements between the two salient poses. Also, compared to the density map using only the landmarks, which cannot accurately identify the salient pose I in the 200-400 and 610-800 frame ranges, the density map obtained using both the landmarks and 5 average joint angles can accurately identify the salient pose I and provide the correct density values $\approx$ 1 for the salient pose I. This sample suggests that integrating the 5 average joint angles with the landmarks in RepCount makes the model more sensitive to salient poses than using only the landmarks, thus improving performance when dealing with the under-counting issue.

\begin{figure}[t!]
    \centering
    \includegraphics[width=\linewidth]{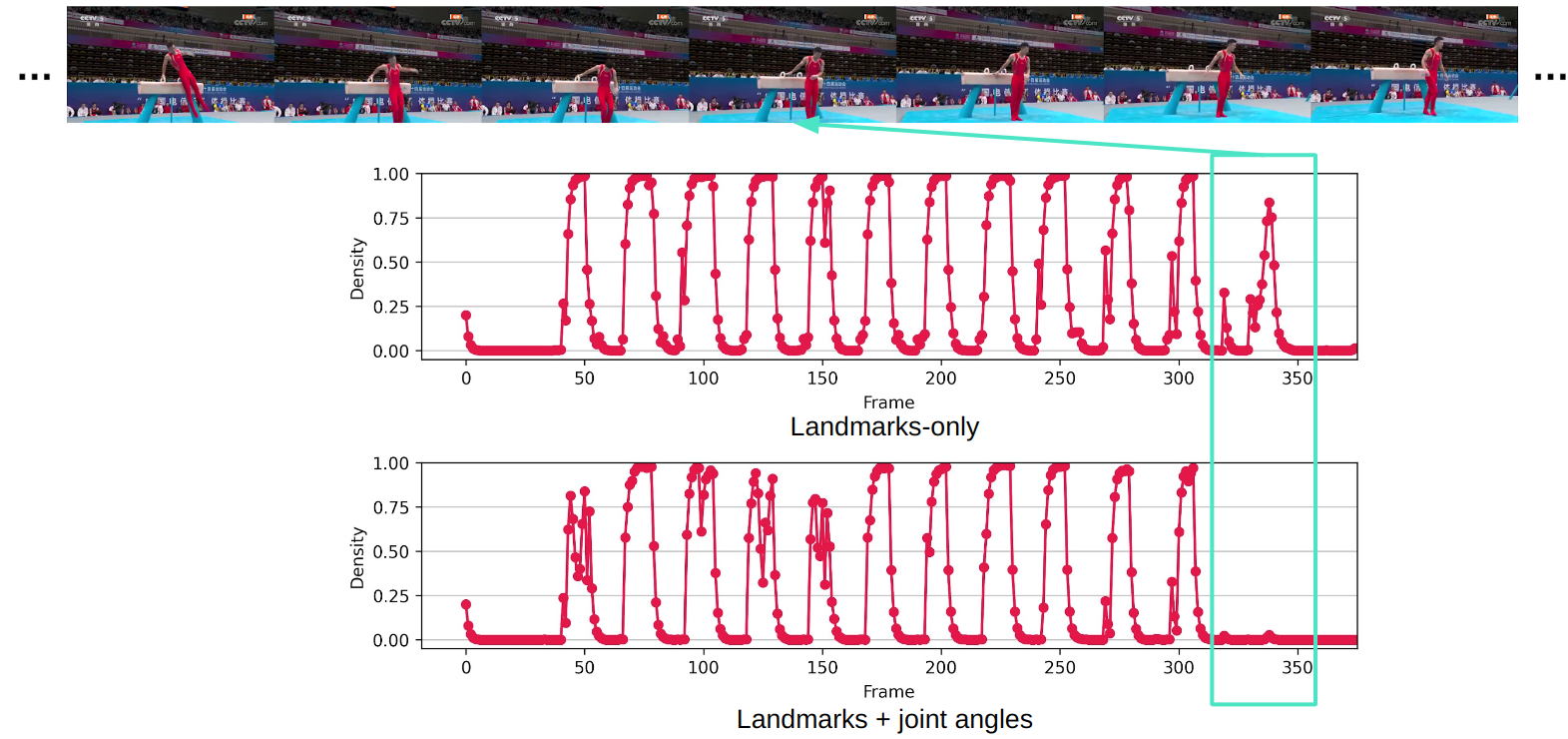}
    \caption{Density maps: landmarks-only vs. landmarks + joint angles - addressing the difficulty in distinguishing sub-actions issue}
    \label{fig:density_sa}
\end{figure}

\textbf{Difficulty in distinguishing sub-actions}: As shown in \cref{fig:density_sa}, an additional jump action is present at the end of the \textit{Pommel Horse} action, which has a lower limb feature similar to the regular \textit{Pommel Horse} action and is a sub-action of the \textit{Pommel Horse}. The density map obtained using only the landmarks treats this jump action as a valid count for \textit{Pommel Horse}, with a peak close to 1 at around frame 340 in \cref{fig:density_sa}, which suggests that an additional count is computed using the landmark-only model, and illustrates the difficulty in distinguishing sub-actions in RepCount using the landmark-only model. However, the density map integrating the 5 average joint angles with the landmarks effectively identifies the jump action is not a valid count for \textit{Pommel Horse}, resulting in a lower density value $\approx$ 0 for this irrelevant sub-action, which in turn provides a more accurate RepCount.

\begin{figure}[t!]
\centering

\begin{subfigure}{\linewidth}
\includegraphics[width=1\linewidth]{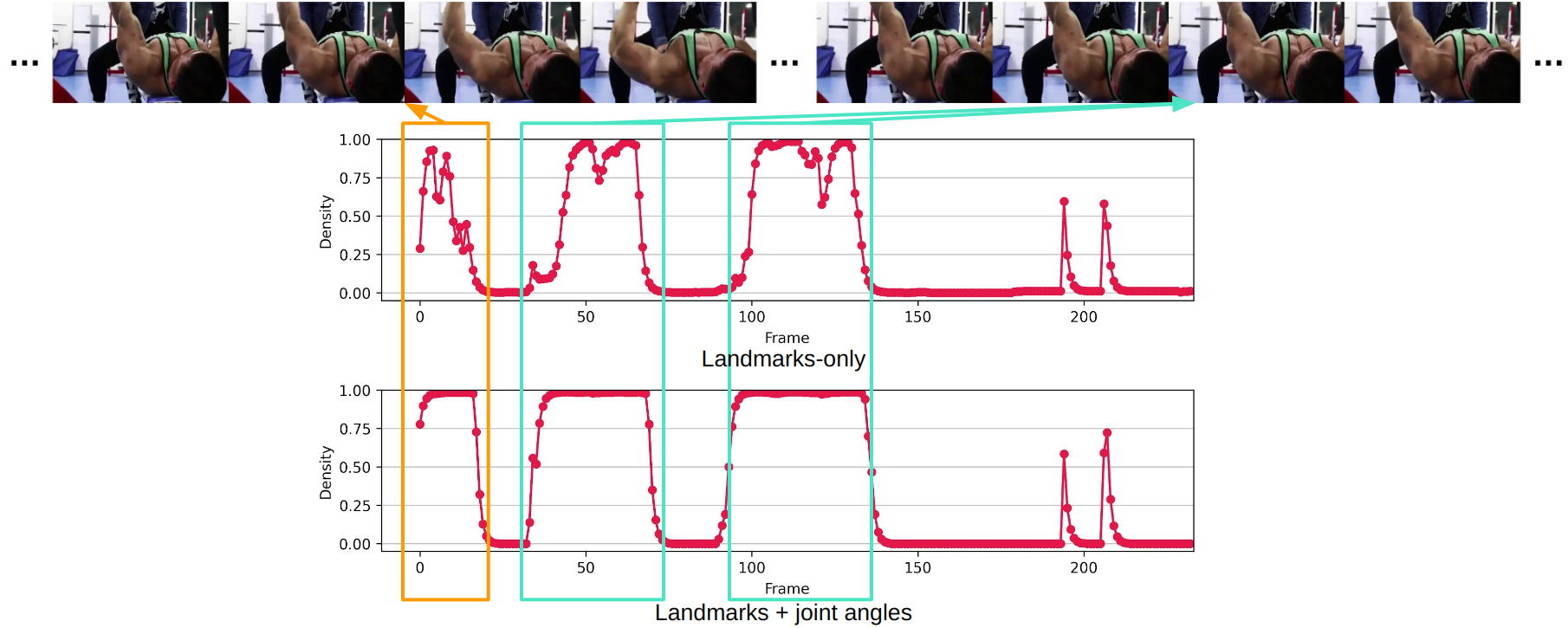} 
\caption{Inaccuracy in recognizing salient poses in \textit{Bench Press}}
\label{fig:density_inac_0}
\end{subfigure}

\begin{subfigure}{\linewidth}
\includegraphics[width=\linewidth]{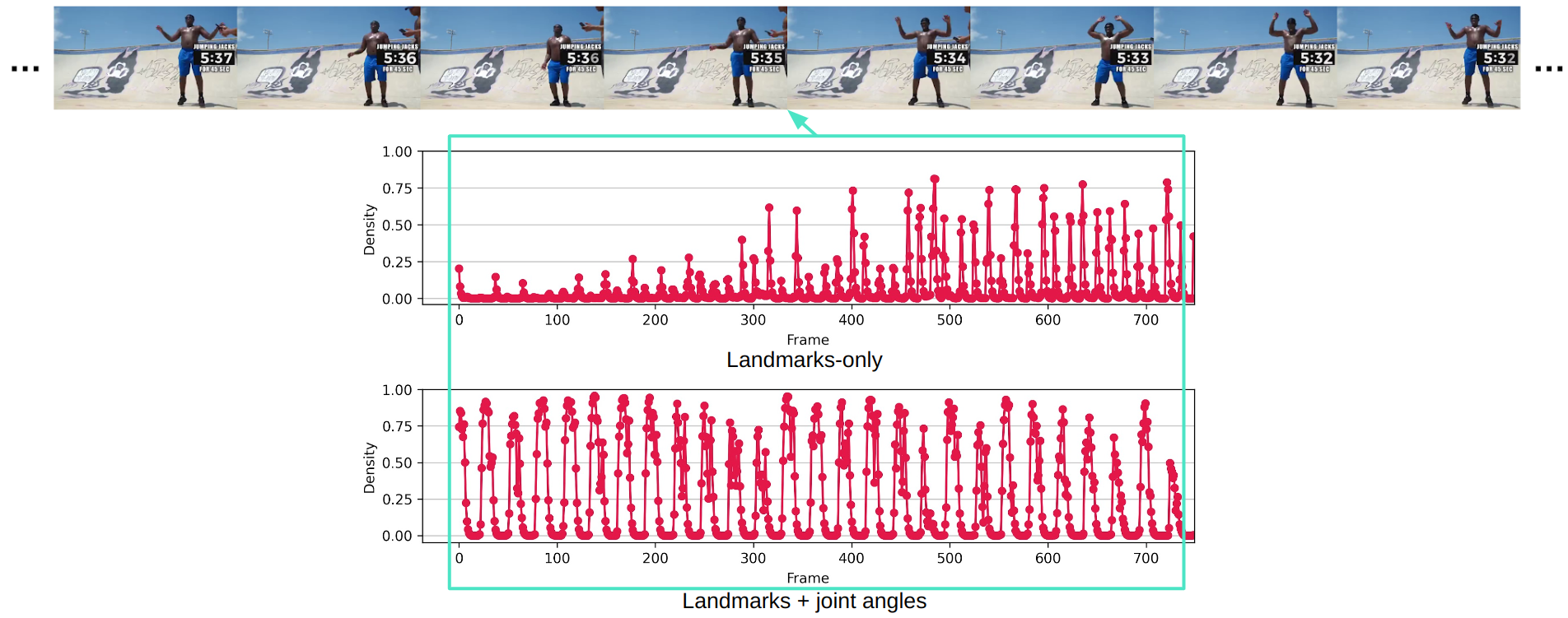}
\caption{Inaccuracy in recognizing salient poses in \textit{Jump Jack}}
\label{fig:density_inac_1}
\end{subfigure}
\caption{Density maps using landmarks-only and landmarks + joint angles  - addressing the inaccuracy issue in recognizing salient poses}
\label{fig:density_inac}
\end{figure}

\textbf{Inaccuracy in recognizing salient poses}: As shown in \cref{fig:density_inac_0}, it is difficult for a model using only the landmarks to provide consistent and reliable density values for the salient pose I (straightened arms) in the RepCount of the action \textit{Bench Press}. Instability occurs when the subject is hindered by factors such as fatigue or slight movements during the execution of the salient pose I, resulting in drops in the density values of the continuous salient pose I, which should be close to 1. However, by integrating the 5 average joint angles with the landmarks, the model consistently and accurately identifies the salient pose I (straightened arms) with a stable density value of 1. A similar performance is shown in \cref{fig:density_inac_1}, where the subject performs the \textit{Jump Jack} action on the right side of the camera view. The density map obtained using only the landmarks does not accurately differentiate between the two salient poses, resulting in small-scale fluctuations between 0 and 0.25 in the beginning half of the density map, while the correct density value for the salient pose I should be close to 1. However, the density map obtained using both the landmarks and 5 average joint angles provides a clear RepCount of the \textit{Jump Jack} action, with the density values fluctuating uniformly between 0 and approximately 1.

These experimental results confirm the advantages of our proposed approach integrating the 5 average joint angles with the landmarks in solving the following issues in RepCount: inability to stably deal with changes in camera viewpoints, over-counting, under-counting, difficulty in distinguishing sub-actions, and inaccuracy in recognizing salient poses, making our method a robust and effective approach for the RepCount task.

\begin{figure}[hbt!]
    \centering
    \includegraphics[width=\linewidth]{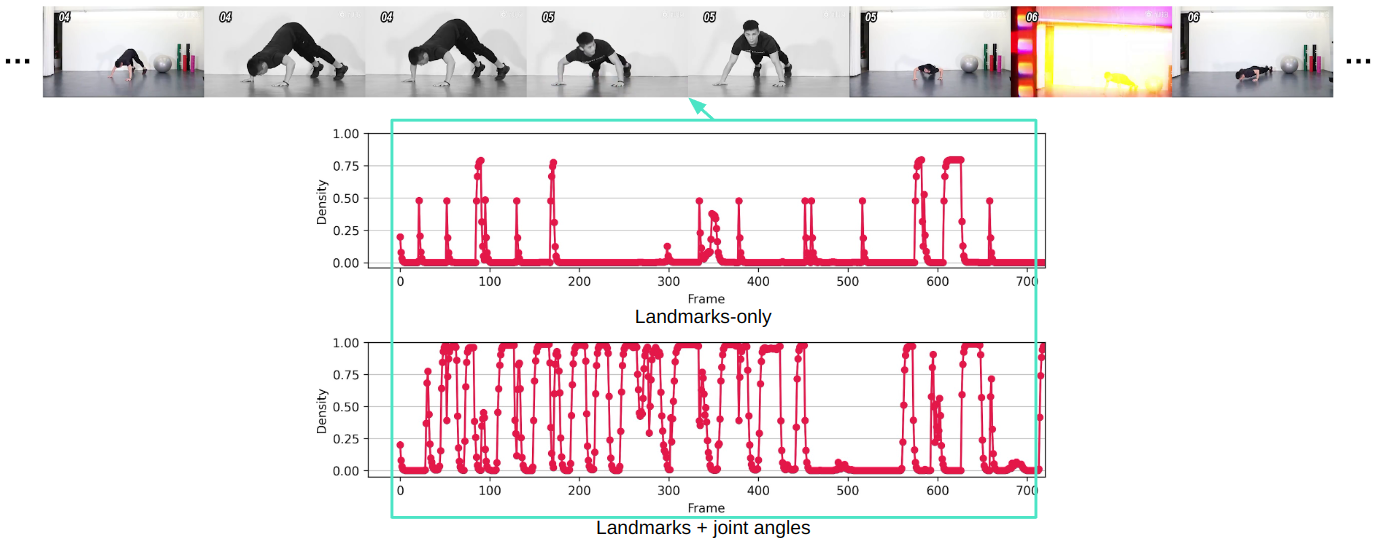}
    \caption{Density maps: landmarks-only vs. landmarks + joint angles - addressing the video effect issue}
    \label{fig:density_ve}
\end{figure}

In addition to the improvements demonstrated above in the regular RepCount cases, we have observed that integrating the 5 average joint angles with the landmarks can lead to effective and robust RepCount in video samples with various video effects, such as instantaneous brightness changes, zoom shifts, etc. \cref{fig:density_ve} illustrates that the density map using both the landmarks and 5 average joint angles provides more accurate results in RepCount than only using the landmarks.

\section{Conclusion}
\label{sec:conclusion}
In summary, this paper integrates the 5 average joint angles and body landmarks in solving the Repetitive Counting (RepCount) problem. Our method significantly improves the performance of RepCount and provides the following improvements: i) Accurate performance in handling camera viewpoint variations. ii) Solving the over-counting and under-counting problems. iii) Improving the recognition of sub-actions. iv) Performing more accurate salient pose recognition. Our method obtains a mean absolute error (MAE) of 0.211 and an off-by-one (OBO) counting accuracy of 0.599. Comprehensive experimental results demonstrate the effectiveness and robustness of our proposed method. Overall, our results outperform previous state-of-the-art PoseRAC methods and point the way to future research in the RepCount problem area.

{\small
\bibliographystyle{unsrt}
\bibliography{egbib}

\begin{thebibliography}{10}

\bibitem{yao2023poserac}
Ziyu Yao, Xuxin Cheng, and Yuexian Zou.
\newblock Poserac: Pose saliency transformer for repetitive action counting.
\newblock {\em arXiv preprint arXiv:2303.08450}, 2023.

\bibitem{hu2022transrac}
Huazhang Hu, Sixun Dong, Yiqun Zhao, Dongze Lian, Zhengxin Li, and Shenghua
  Gao.
\newblock Transrac: Encoding multi-scale temporal correlation with transformers
  for repetitive action counting.
\newblock In {\em Proceedings of the IEEE/CVF Conference on Computer Vision and
  Pattern Recognition}, pages 19013--19022, 2022.

\bibitem{albu2008generic}
A~Branzan Albu, Robert Bergevin, and S{\'e}bastien Quirion.
\newblock Generic temporal segmentation of cyclic human motion.
\newblock {\em Pattern Recognition}, 41(1):6--21, 2008.

\bibitem{chen2023real}
Haodong Chen, Niloofar Zendehdel, Ming~C Leu, and Zhaozheng Yin.
\newblock Real-time human-computer interaction using eye gazes.
\newblock {\em Manufacturing Letters}, 35:883--894, 2023.

\bibitem{gonzalez2021durable}
Andrew Gonzalez, Hamada~A Aboubakr, John Brockgreitens, Weixing Hao, Yang Wang,
  Sagar~M Goyal, and Abdennour Abbas.
\newblock Durable nanocomposite face masks with high particulate filtration and
  rapid inactivation of coronaviruses.
\newblock {\em Scientific reports}, 11(1):24318, 2021.

\bibitem{briassouli2007extraction}
Alexia Briassouli and Narendra Ahuja.
\newblock Extraction and analysis of multiple periodic motions in video
  sequences.
\newblock {\em IEEE transactions on pattern analysis and machine intelligence},
  29(7):1244--1261, 2007.

\bibitem{hao2020filtration}
Weixing Hao, Andrew Parasch, Stephen Williams, Jiayu Li, Hongyan Ma, Joel
  Burken, and Yang Wang.
\newblock Filtration performances of non-medical materials as candidates for
  manufacturing facemasks and respirators.
\newblock {\em International journal of hygiene and environmental health},
  229:113582, 2020.

\bibitem{chen2020design}
Hao-dong Chen, Hongbo Zhu, Zhiqiang Teng, and Ping Zhao.
\newblock Design of a robotic rehabilitation system for mild cognitive
  impairment based on computer vision.
\newblock {\em Journal of Engineering and Science in Medical Diagnostics and
  Therapy}, 3(2):021108, 2020.

\bibitem{brickwood2019consumer}
Katie-Jane Brickwood, Greig Watson, Jane O'Brien, Andrew~D Williams, et~al.
\newblock Consumer-based wearable activity trackers increase physical activity
  participation: systematic review and meta-analysis.
\newblock {\em JMIR mHealth and uHealth}, 7(4):e11819, 2019.

\bibitem{foran2001high}
Bill Foran.
\newblock {\em High-performance sports conditioning}.
\newblock Human kinetics, 2001.

\bibitem{hao2021factors}
Weixing Hao, Guang Xu, and Yang Wang.
\newblock Factors influencing the filtration performance of homemade face
  masks.
\newblock {\em Journal of Occupational and Environmental Hygiene},
  18(3):128--138, 2021.

\bibitem{shen2017milift}
Chenguang Shen, Bo-Jhang Ho, and Mani Srivastava.
\newblock Milift: Efficient smartwatch-based workout tracking using automatic
  segmentation.
\newblock {\em IEEE Transactions on Mobile Computing}, 17(7):1609--1622, 2017.

\bibitem{chen2022real}
Haodong Chen, Ming~C Leu, and Zhaozheng Yin.
\newblock Real-time multi-modal human--robot collaboration using gestures and
  speech.
\newblock {\em Journal of Manufacturing Science and Engineering},
  144(10):101007, 2022.

\bibitem{dwibedi2020counting}
Debidatta Dwibedi, Yusuf Aytar, Jonathan Tompson, Pierre Sermanet, and Andrew
  Zisserman.
\newblock Counting out time: Class agnostic video repetition counting in the
  wild.
\newblock In {\em Proceedings of the IEEE/CVF conference on computer vision and
  pattern recognition}, pages 10387--10396, 2020.

\bibitem{zhang2020context}
Huaidong Zhang, Xuemiao Xu, Guoqiang Han, and Shengfeng He.
\newblock Context-aware and scale-insensitive temporal repetition counting.
\newblock In {\em Proceedings of the IEEE/CVF Conference on Computer Vision and
  Pattern Recognition}, pages 670--678, 2020.

\bibitem{bazarevsky2020blazepose}
Valentin Bazarevsky, Ivan Grishchenko, Karthik Raveendran, Tyler Zhu, Fan
  Zhang, and Matthias Grundmann.
\newblock Blazepose: On-device real-time body pose tracking.
\newblock {\em arXiv preprint arXiv:2006.10204}, 2020.

\bibitem{chen2023fine}
Haodong Chen, Niloofar Zendehdel, Ming~C Leu, and Zhaozheng Yin.
\newblock Fine-grained activity classification in assembly based on
  multi-visual modalities.
\newblock {\em Journal of Intelligent Manufacturing}, pages 1--19, 2023.

\bibitem{liu2022video}
Ze~Liu, Jia Ning, Yue Cao, Yixuan Wei, Zheng Zhang, Stephen Lin, and Han Hu.
\newblock Video swin transformer.
\newblock In {\em Proceedings of the IEEE/CVF conference on computer vision and
  pattern recognition}, pages 3202--3211, 2022.

\bibitem{onoro2016towards}
Daniel Onoro-Rubio and Roberto~J L{\'o}pez-Sastre.
\newblock Towards perspective-free object counting with deep learning.
\newblock In {\em Computer Vision--ECCV 2016: 14th European Conference,
  Amsterdam, The Netherlands, October 11--14, 2016, Proceedings, Part VII 14},
  pages 615--629. Springer, 2016.

\bibitem{sreenu2019intelligent}
G~Sreenu and Saleem Durai.
\newblock Intelligent video surveillance: a review through deep learning
  techniques for crowd analysis.
\newblock {\em Journal of Big Data}, 6(1):1--27, 2019.

\bibitem{huang2020improving}
Yifei Huang, Yusuke Sugano, and Yoichi Sato.
\newblock Improving action segmentation via graph-based temporal reasoning.
\newblock In {\em Proceedings of the IEEE/CVF conference on computer vision and
  pattern recognition}, pages 14024--14034, 2020.

\end{thebibliography}
}

\end{document}